%% file: main.tex
\documentclass[sigconf,nonacm]{acmart}
\usepackage{booktabs}
\usepackage{graphicx}
\usepackage{tabularx}
\usepackage{pifont}
\newcommand{\cmark}{\ding{51}}
\setcopyright{none}
\renewcommand\footnotetextcopyrightpermission[1]{}
\begin{document}
\title[NEIGHBOR: Service Behaviour and Evidence Use]{NEIGHBOR: Aligning Service Behaviour and Evidence Use in an LLM Agent for Community Resident Services}
\author{Zongyou Yang}
\affiliation{%
  \department{Dyson School of Design Engineering}
  \institution{Imperial College London}
  \city{London}
  \country{United Kingdom}}
\email{zy2926@ic.ac.uk}
\author{Yinghan Hou}
\affiliation{%
  \department{Department of Electrical and Electronic Engineering}
  \institution{Imperial College London}
  \city{London}
  \country{United Kingdom}}
\email{yh24@ic.ac.uk}
\author{Xiaokun Yang}
\authornote{Corresponding author.}
\affiliation{%
  \department{School of Electronic Information}
  \institution{Nanchang Institute of Technology}
  \city{Nanchang}
  \country{China}}
\email{yangxk@bupt.cn}

\begin{abstract}
Residents who contact a community service desk about a repair or a service charge need answers they can act on, with facts that hold for their case and commitments the desk can keep. In tool-augmented LLM agents, such facts come from documents and records, and the conduct of a service answer is usually specified in a system prompt. We present NEIGHBOR, an LLM agent for community resident services that trains this conduct into the model that drives its tools. NEIGHBOR is deployed on a live resident-service platform for a district of an East Asian city of more than ten million people. It adapts Qwen3-235B-A22B with LoRA on 7,824 supervised fine-tuning examples built from public community-service materials around six service-behaviour targets, among them commitment control and human referral, and the adapted model plans document retrieval and read-only text-to-SQL over the resident's own records. In an offline evaluation with seven configurations, an independent domain expert rated all 350 answers to 50 stratified held-out queries. NEIGHBOR scores 7.74/10 against 5.86 for the same agent running the un-tuned model (+1.88; paired Wilcoxon $p<10^{-8}$; higher on 46 of 50 queries) and improves all six rated dimensions. No tool-augmented answer received a red-line flag (standalone models: 8--14\% of answers). An LLM judge ranks the seven configurations much as the expert does (Spearman $\rho=0.93$). Its rubric counts specifics absent from the reference answer as fabricated, so it scores tool-grounded answers 1--2 points lower, enough to change which configuration ranks first. For resident-facing agents, service behaviour is a practical fine-tuning target that raises expert-rated answer quality with the tools held fixed.
\end{abstract}

\maketitle
\raggedbottom
\input{body}
\balance
\bibliographystyle{ACM-Reference-Format}
\bibliography{references}
\end{document}

%% file: body.tex
\section{Introduction}
Community service desks handle resident requests that range from a leaking pipe or a parking permit to a service charge that looks wrong. Large language model (LLM) assistants can take on much of this front-line work when their answers are right for the resident's case and follow the conduct of an experienced service desk. A resident who questions a service charge, for example, needs the balance on record and a clear route to the staff who can review it.

Retrieval-augmented generation (RAG) and tool use ground answers in documents and records~\cite{lewis2020rag,gao2024ragsurvey,yao2022react}, and service conduct is usually specified in a system prompt. Experienced service staff learn this conduct by example, from acknowledging a complaint to recognising when a request needs someone with authority. These conventions recur from case to case, which makes them a good target for supervised fine-tuning (SFT).

\begin{figure*}[t]
\centering
\includegraphics[width=\textwidth]{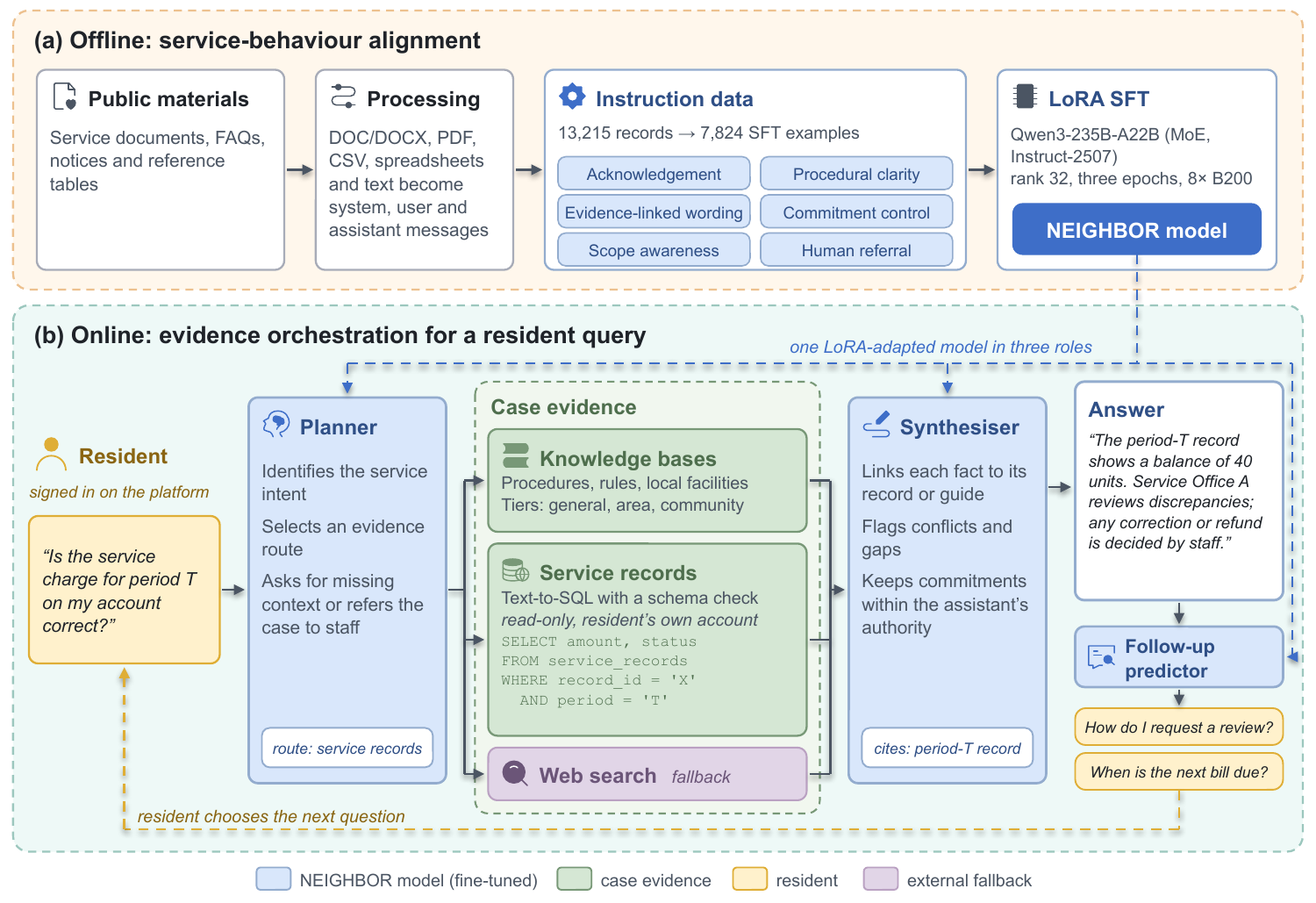}
\Description{Two-panel overview diagram. Panel (a), offline: public materials, processing, instruction data with six behaviour-target labels, and LoRA SFT of Qwen3-235B-A22B, which produces the NEIGHBOR model. Panel (b), online: a resident question about a service charge goes to the planner, then to case evidence (knowledge bases, service records queried through text-to-SQL with a schema check, and a web-search fallback), then to the synthesiser and an answer that cites the period-T record. A follow-up predictor suggests two questions, and a dashed yellow path returns to the resident. A dashed blue line links the NEIGHBOR model to its three roles.}
\caption{NEIGHBOR overview, traced with the synthetic billing example of Section~\ref{sec:behaviour}. (a)~Offline, public service materials become instruction data written to six behaviour targets, and LoRA SFT adapts Qwen3-235B-A22B into the NEIGHBOR model. (b)~Online, the same model plans an evidence route, synthesises an answer that cites the period-T record, and proposes follow-up questions, from which the resident picks the next one to ask. Blue marks the fine-tuned model, green the case evidence, yellow the resident, and lavender the web-search fallback.}
\label{fig:architecture}
\end{figure*}

NEIGHBOR trains this conduct, including evidence-linked wording, into one adapted model and connects that model to the documents and records that hold the facts of each case. We adapt Qwen3-235B-A22B~\cite{qwen2025qwen3} with LoRA~\cite{hu2021lora} on instruction data built from public community-service materials and organised around six service-behaviour targets (Table~\ref{tab:behaviour}). At run time, the adapted model plans an evidence route over tiered knowledge bases and schema-checked text-to-SQL over service records, with web search as a fallback, then writes an evidence-linked answer with bounded commitments and suggests follow-up questions (Figure~\ref{fig:architecture}).

NEIGHBOR is deployed on a live resident-service platform that serves a district of an East Asian city with more than ten million residents. This paper reports an offline system case study in that setting: seven configurations answer 300 held-out resident queries, and an independent domain expert rates all seven answers to a stratified subset of 50 queries on six dimensions. We obtain four main results.
\begin{itemize}
\item With the tools held fixed, fine-tuning raises the agent's expert score from 5.86 to 7.74/10 (+1.88; paired Wilcoxon $p<10^{-8}$, $d_z=1.47$), more than twice the 0.76 points that the agent and its tools add to the standalone Qwen3-235B.
\item NEIGHBOR improves all six rated dimensions and has the highest mean of the seven configurations on each. Utility (+2.76), retrieval (+2.60), and empathy (+2.52) gain most, and with the tools unchanged, the retrieval gain means the fine-tuned model's answers cover more of the key information.
\item None of the 200 tool-augmented answers drew a red-line flag (standalone models: 8--14\%), and NEIGHBOR-Full, which combines documents with records, scores above both single-channel variants.
\item An LLM judge ranks the seven configurations much as the expert does (Spearman $\rho=0.93$) and correlates with the expert on individual answers ($\rho=0.556$). Its rubric counts specifics absent from the reference answer as fabricated, so it scores tool-grounded answers 1--2 points lower, mostly on hallucination control, enough to rank the document-only variant above the full system.
\end{itemize}
This paper contributes six service-behaviour targets and the instruction data built around them (Section~\ref{sec:behaviour}), the NEIGHBOR agent, which runs one adapted model over document and record tools (Section~\ref{sec:system}), and a seven-configuration expert evaluation with a judge--expert comparison (Sections~\ref{sec:evaluation} and~\ref{sec:results}), from which we derive design principles for agency-preserving resident agents (Section~\ref{sec:design}).

\section{Related Work}
\paragraph{Instruction tuning for domain conduct.}
Instruction tuning and preference alignment shape how language models respond to requests~\cite{wei2022finetuned,ouyang2022instruct}, and a small, carefully curated set of examples can set a model's response style~\cite{zhou2023lima}. Self-Instruct and data-centric practice emphasise how supervision is constructed and curated~\cite{wang2023selfinstruct,zha2023datacentric,whang2023dataquality}, and adapters, LoRA, and QLoRA make domain adaptation of large models affordable~\cite{houlsby2019peft,hu2021lora,dettmers2023qlora}. Empathetic-dialogue data train models to respond to the feelings a speaker describes~\cite{rashkin2019empathetic}. Domain adaptation typically targets task competence and domain knowledge, and NEIGHBOR extends it to service conduct through the six behaviour targets of Section~\ref{sec:behaviour}.

\paragraph{Tool-grounded agents.}
RAG supplies external evidence at generation time~\cite{lewis2020rag,gao2024ragsurvey}, building on dense, lexical, and approximate nearest-neighbour retrieval~\cite{karpukhin2020dpr,robertson2009bm25,malkov2020hnsw}. Retrieval reduces hallucination in knowledge-grounded conversation~\cite{shuster2021retrieval}, and Self-RAG learns when to retrieve and how to critique its own output~\cite{asai2024selfrag}. ReAct interleaves reasoning with tool calls~\cite{yao2022react}, Toolformer learns through self-supervision when to call an API~\cite{schick2024toolformer}, ToolLLM trains models on more than 16,000 real-world APIs~\cite{qin2024toolllm}, and agent surveys place tools alongside planning and memory~\cite{wang2024survey,xu2026agent}. Text-to-SQL extends grounding to structured records~\cite{yu2018spider,li2023dawn}. Tool-using agents also follow system-side instructions that specify tools, policies, and protocols, and compressing these instructions shifts their reliability in ways that depend on the method and the share retained~\cite{hou2026control}. NEIGHBOR trains service conduct into the model itself and measures what this adds over document and database grounding when the tools stay fixed.

\paragraph{Human--AI interaction and evaluation.}
Human-centred AI asks for systems that are reliable, safe, and trustworthy~\cite{shneiderman2020hcai}, and appropriate reliance on automation depends on whether people can tell when to trust it~\cite{lee2004trust}. Chatbots and other AI in the public sector bring these questions into citizen services~\cite{androutsopoulou2019chatbots,wirtz2019public}. Guidelines for human--AI interaction, which inform NEIGHBOR's behaviour targets, ask systems to make clear what they can do, scope services when in doubt, and support efficient correction~\cite{amershi2019guidelines}. Mixed-initiative design coordinates automated assistance with user control~\cite{horvitz1999mixed}. LLM judges and crowdsourced comparison platforms scale response assessment~\cite{zheng2023judging,liu2023geval,chiang2024chatbot}, and the position, identity, and protocol of the judge shape what is measured~\cite{wang2024fair,chen2025selfpref,yang2026judgechanges}. Factuality evaluation checks generated claims against a knowledge source~\cite{ji2023hallucination,min2023factscore}, so what counts as supported depends on that source, and aggregate accuracy can hide how models fail~\cite{yang2026accuracy}. In our comparison of an LLM judge with a domain expert, the two rank the configurations alike, and this agreement hides a divergence on answers whose tools supply evidence that the reference answer does not contain.

\section{Service Behaviour as a Training Target}
\label{sec:behaviour}
\subsection{Six Behaviour Targets}
We organise supervision around six behaviour targets, each of which constrains how an answer serves the resident (Table~\ref{tab:behaviour}).

\begin{table}[tbp]
\centering\small
\caption{Service-behaviour targets and the response constraint each one encodes.}
\label{tab:behaviour}
\begin{tabularx}{\columnwidth}{@{}p{0.30\columnwidth}X@{}}
\toprule
Target & Response constraint \\
\midrule
Acknowledgement & Recognise the request and the resident's concern without inventing details about their situation. \\
Procedural clarity & Give the next step as an ordered, actionable sequence. \\
Evidence-linked wording & Attribute facts to the supplied record or guide, separately from general advice. \\
Commitment control & Keep promises about acceptance, corrections, refunds, and completion within the assistant's authority. \\
Scope awareness & Identify missing context and requests beyond the available service information. \\
Human referral & Direct the resident to the right staff channel when evidence or authority is insufficient. \\
\bottomrule
\end{tabularx}
\end{table}

The targets work together on uncertain cases, as a synthetic billing example shows. The record for Community~A and Account~X lists a charge of 120 units, a payment of 80, and a balance of 40 for period~T, and the resident asks whether the charge is correct. The target answer acknowledges the concern and states that the period-T record shows a balance of 40. It then names Service Office~A as the channel for reviewing a discrepancy and leaves any correction or refund to staff.

\subsection{Instruction Data from Public Service Materials}
NEIGHBOR serves residents of an urban district in an East Asian city of more than ten million people, and its materials and queries are in Chinese. We collect publicly accessible service documents, FAQs, notices, and reference tables, and processing scripts convert DOC/DOCX, PDF, CSV, spreadsheet, and plain-text sources into system, user, and assistant messages. Documents are segmented into coherent passages, and table rows keep their field meanings. Questions are phrased the way residents ask them, including colloquial wording and vague references, and reference answers pair the factual content of the source with the conduct set out in the behaviour targets.

\begin{figure*}[t]
\centering
\includegraphics[width=\textwidth]{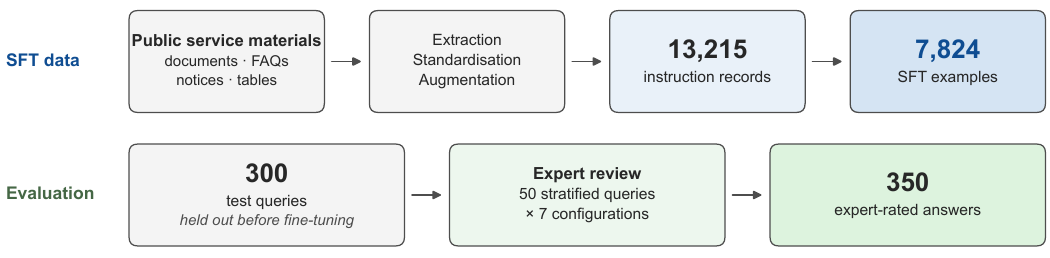}
\Description{Left-to-right flow: public service materials are extracted, standardised and augmented into 13,215 instruction records, of which 7,824 form the SFT dataset; a separate held-out branch leads from 300 test queries to the expert review of 50 stratified queries answered by seven configurations, 350 rated answers in total.}
\caption{Data and evaluation design. Public service materials become 13,215 instruction records and a 7,824-example SFT dataset. The 300 held-out test queries feed the seven-configuration comparison, and the expert rates all seven answers to a stratified 50-query subset.}
\label{fig:dataset}
\end{figure*}

Collection, standardisation, and augmentation yield 13,215 instruction records. The 7,824-example SFT dataset is selected from these records, and the 300 held-out test queries are drawn from the records outside it, so none of them is used for training or checkpoint selection (Figure~\ref{fig:dataset}). The 50 queries for expert review form a stratified subset of these 300 (Section~\ref{sec:evaluation}).

\subsection{Adaptation}
For request $q$, context $c$, and reference response $r$, adaptation minimises the standard assistant-token objective
\begin{equation}
\mathcal{L}_{\mathrm{SFT}}(\theta)=-\sum_t\log P_{\theta}(r_t\mid q,c,r_{<t}).
\end{equation}
We adapt the mixture-of-experts Qwen3-235B-A22B (Instruct-2507) with LoRA on eight NVIDIA B200 GPUs: rank 32, alpha 32, dropout 0.1, three epochs, global batch size 16, maximum sequence length 32,768, learning rate $10^{-6}$ with cosine scheduling and warmup fraction 0.03, weight decay 0.01, packing disabled, validation every 16 steps, and checkpoints every 64 steps.

\section{Evidence Orchestration}
\label{sec:system}
Figure~\ref{fig:architecture} shows the offline adaptation path and the online workflow, in which one LoRA-adapted model acts as planner, synthesiser, and follow-up predictor. The planner selects the evidence a request needs, and the synthesiser writes the answer from the tool results in the service register learned from the instruction data.

\paragraph{Planning.} From the request and its conversation context, the planner identifies the service intent and selects an evidence route. Record lookups such as balances, payments, and service histories go to the database, and questions about procedures, rules, and local facilities go to document retrieval. Web search covers requests outside local coverage, and the synthesiser presents its results as external material. Requests that need clarification or staff authority receive a clarifying question or a referral.

\paragraph{Document retrieval.} The knowledge bases separate general service guides, area-level information, and community-level materials, and they keep current and historical versions apart, so retrieval can prefer the most specific and current source.

\paragraph{Schema-checked database access.} Record queries are available to residents who have signed in on the platform. The model generates SQL, a validator checks table and field names against the schema, and validated queries run through a read-only database account and are restricted to the signed-in resident's own records, so the agent reads no other account and changes no data. In the billing example, the generic fields \texttt{community\_\allowbreak alias}, \texttt{record\_id}, \texttt{service\_type}, \texttt{period}, \texttt{amount}, and \texttt{status} return the balance that the synthesiser explains (Figure~\ref{fig:architecture}b).

\paragraph{Synthesis and follow-up.} The synthesiser receives the request and all tool results. It states which record or guide supports each fact, flags conflicts and gaps, and phrases commitments within the assistant's authority. A follow-up call proposes two related questions, which the resident can pick as the next query. Payments, record changes, and maintenance actions remain staff actions, reached through referral. All roles call the same adapted model, served privately with vLLM~\cite{kwon2023vllm}.

\section{Evaluation Setup}
\label{sec:evaluation}
\subsection{Configurations}
Seven configurations answer the 300 held-out queries (Table~\ref{tab:configurations}). Three standalone models (Qwen3-8B, Qwen3-235B, and ERNIE-4.5-21B-A3B from a different model family) answer without local tools. The agent baseline runs the agent with both evidence channels and the un-tuned Qwen3-235B model, so its comparison with NEIGHBOR-Full measures what fine-tuning adds with the tools held fixed. NEIGHBOR-RAG and NEIGHBOR-DB keep the fine-tuned model and remove one evidence channel each.

\begin{table}[tbp]
\centering\small\setlength{\tabcolsep}{4pt}
\caption{Evaluated configurations: domain SFT and local evidence channels.}
\label{tab:configurations}
\begin{tabular}{@{}lccc@{}}
\toprule
Configuration & SFT & Documents & Database \\
\midrule
Qwen3-8B (standalone) & -- & -- & -- \\
Qwen3-235B (standalone) & -- & -- & -- \\
ERNIE-4.5-21B-A3B (standalone) & -- & -- & -- \\
Agent baseline (Qwen3-235B) & -- & \cmark & \cmark \\
NEIGHBOR-RAG & \cmark & \cmark & -- \\
NEIGHBOR-DB & \cmark & -- & \cmark \\
NEIGHBOR-Full & \cmark & \cmark & \cmark \\
\bottomrule
\end{tabular}
\end{table}

\subsection{Test Queries and Expert Review}
The 300 held-out queries reflect what residents ask about: administrative procedures (40\%), community affairs (12\%), contact lookups (12\%), service charges (11\%), general enquiries (9\%), and complaint, policy, social-security, and property-service questions (15\%). Their sources range from community-level materials (40\%) through city-level materials (30\%) to provincial and national materials (15\% each). Simple retrievals make up 80\% of the queries, and 20\% require multi-step reasoning. A stratified subset of 50 queries preserves this composition: 42 simple and eight multi-step queries, drawn 21, 15, 7, and 7 from community, city, provincial, and national materials. An independent domain expert rates all seven answers to each of the 50 queries, giving 350 rated answers.

\subsection{Scoring}
Each answer is rated from 1 to 10 on six dimensions: empathy $E$ (acknowledging the resident's concern before resolving it), utility $U$ (actionable steps, offices, and materials), compliance $B$ (no commitment beyond the assistant's authority, with out-of-scope requests redirected), hallucination control $H$ (no fabricated or contradicted facts), conciseness $C$, and retrieval $R$ (coverage of key information such as fees, contacts, and clauses). The weighted score emphasises factual reliability:
\begin{equation}
S=0.10E+0.20U+0.15B+0.25H+0.20C+0.10R.
\end{equation}
A red-line flag, raised for fabricated policy clauses or contact details, commitments beyond the assistant's authority, or privacy violations, caps the score at 1. We analyse raw weighted scores and report red-line-capped means alongside them, and the two coincide for every tool-augmented configuration.

Both assessors receive the question, its reference answer, and the candidate answer, and both apply these dimensions, weights, and red-line rules. Retrieved passages and database records are not supplied as separate material, so tool evidence reaches an assessor through what the answer states and cites. The expert scores the seven answers to a question together, ranks them from 1 to 7 before assigning absolute scores, and records a rationale for each answer. The answers carry shuffled letters in place of system names, and the expert does not see the LLM judge's scores. Four principles guide these scores. Specific facts such as contacts, addresses, and policies are checked against the reference answer: consistent core information earns credit, contradictions are penalised, and reasonable additions are not. Citation markers that point to retrieved passages do not count as fabrication. Tool use earns no credit by itself: retrieved content counts when it is consistent with the reference answer, relevant, and used appropriately, and a conflict between knowledge-base and database results that the answer leaves unresolved lowers hallucination control and compliance. Red-line thresholds are the same for every configuration.

\subsection{Automated Judge and Statistics}
The same 350 answers are also scored by an LLM judge, MiniMax-M2.7, which receives the query, the system role, the reference answer, and the candidate answer. Its rubric defines five score bands per dimension, treats the reference answer as the factual benchmark, and counts a specific contact, address, clause, or amount as fabricated when the reference does not contain it and the question does not imply it. Paired comparisons use the question as the unit of analysis: Wilcoxon signed-rank tests over the 50 questions, Cohen's $d_z$, and 95\% $t$-intervals of the mean difference.

\section{Results}
\label{sec:results}
\subsection{Fine-Tuning Adds 1.88 Points on Top of Identical Tools}
\begin{figure}[tbp]
\centering
\includegraphics[width=\columnwidth]{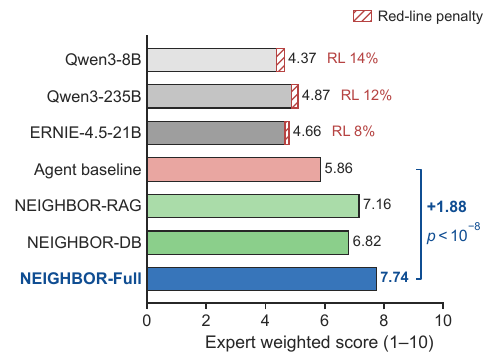}
\Description{Horizontal bar chart of expert weighted scores for seven configurations, with hatched red-line penalties for the three standalone models and a bracket marking the 1.88-point gap between the agent baseline and NEIGHBOR-Full.}
\caption{Expert weighted scores after the red-line cap. Hatched segments show the red-line penalty and RL the red-line rate; both fall only on standalone models. NEIGHBOR-Full leads the agent baseline that shares its tools by 1.88 points (paired Wilcoxon, $n=50$).}
\label{fig:overall}
\end{figure}

With the evidence channels held fixed, fine-tuning raises the expert score from 5.86 to 7.74, the highest of the seven configurations (Table~\ref{tab:results}, Figure~\ref{fig:overall}). The agent baseline and NEIGHBOR-Full call the same tools, so the 1.88-point gain (95\% CI 1.51--2.24; Wilcoxon $p=3.4\times10^{-9}$; $d_z=1.47$) is what fine-tuning adds, and it holds on 46 of the 50 queries, with one tie. The SFT data pair the behaviour targets with the factual content of the source materials (Section~\ref{sec:behaviour}), so the gain measures the fine-tuning step as a whole, and Section~\ref{sec:dims} locates it by dimension. NEIGHBOR-Full also leads each standalone model, Qwen3-235B by 2.64 points, ERNIE-4.5-21B by 2.94, and Qwen3-8B by 3.11 (all $p<10^{-8}$), and it scores higher than ERNIE-4.5-21B and Qwen3-8B on every one of the 50 queries.

\begin{table}[tbp]
\centering\small\setlength{\tabcolsep}{3.2pt}
\caption{Expert review of 50 queries: mean weighted score (raw and red-line capped), median, share of answers scoring at least 7, and red-line rate.}
\label{tab:results}
\begin{tabular}{@{}lrrrrr@{}}
\toprule
System & Raw & Capped & Median & $\geq$7 & RL \\
\midrule
Qwen3-8B & 4.633 & 4.367 & 4.65 & 0\% & 14\% \\
Qwen3-235B & 5.100 & 4.867 & 5.25 & 2\% & 12\% \\
ERNIE-4.5-21B & 4.797 & 4.657 & 4.78 & 0\% & 8\% \\
Agent baseline & 5.861 & 5.861 & 5.40 & 14\% & 0\% \\
NEIGHBOR-RAG & 7.159 & 7.159 & 6.78 & 48\% & 0\% \\
NEIGHBOR-DB & 6.816 & 6.816 & 6.88 & 46\% & 0\% \\
\textbf{NEIGHBOR-Full} & \textbf{7.740} & \textbf{7.740} & \textbf{8.10} & \textbf{78\%} & \textbf{0\%} \\
\bottomrule
\end{tabular}
\end{table}

From the standalone Qwen3-235B to NEIGHBOR-Full, the score rises in two cumulative steps. Placing the un-tuned model inside the agent with both evidence channels adds 0.76 points ($p=0.005$), and fine-tuning the model inside the same agent adds a further 1.88, more than twice the first step.

\subsection{Gains on Every Rated Dimension}
\label{sec:dims}
\begin{figure*}[tbp]
\centering
\includegraphics[width=\textwidth]{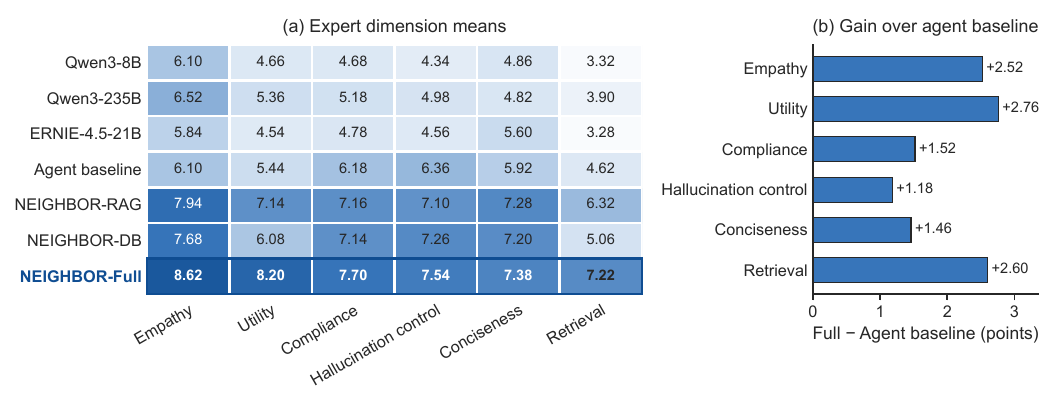}
\Description{A heatmap of expert dimension means for seven configurations next to a bar chart of the per-dimension gain of NEIGHBOR-Full over the agent baseline.}
\caption{Where the gain occurs. (a)~Expert dimension means; NEIGHBOR-Full has the highest mean on every dimension. (b)~Gain of NEIGHBOR-Full over the agent baseline, in rating points (each $p<0.001$, paired Wilcoxon, $n=50$).}
\label{fig:profiles}
\end{figure*}

NEIGHBOR-Full improves on the agent baseline in all six dimensions (Figure~\ref{fig:profiles}), each with $p<0.001$: utility by 2.76 points, retrieval by 2.60, empathy by 2.52, compliance by 1.52, conciseness by 1.46, and hallucination control by 1.18. It has the highest mean of all seven configurations on every dimension, and its empathy score improves on 49 of the 50 queries and never declines.

The dimension profile matches the behaviour targets. The utility and empathy gains correspond to procedural clarity and acknowledgement, and the compliance gain to commitment control. With the same tools available, the fine-tuned model's answers also cover more of the key information and contradict fewer facts, in line with the evidence-linked wording target.

\subsection{Red-Line Outcomes and Combined Evidence Channels}
None of the 200 answers from the four tool-augmented configurations received a red-line flag. The expert flagged 7, 6, and 4 of the 50 answers from the standalone Qwen3-8B, Qwen3-235B, and ERNIE-4.5-21B (Table~\ref{tab:results}). Among the fine-tuned variants, NEIGHBOR-Full, which draws on both evidence channels, exceeds the document-only variant by 0.58 points ($p=1.5\times10^{-4}$) and the database-only variant by 0.92 ($p=3.9\times10^{-4}$). With both channels available, the planner routes record lookups to the database and questions about procedures and rules to the documents (Section~\ref{sec:system}).

\subsection{Robustness of the Gain}
\begin{figure*}[tbp]
\centering
\includegraphics[width=\textwidth]{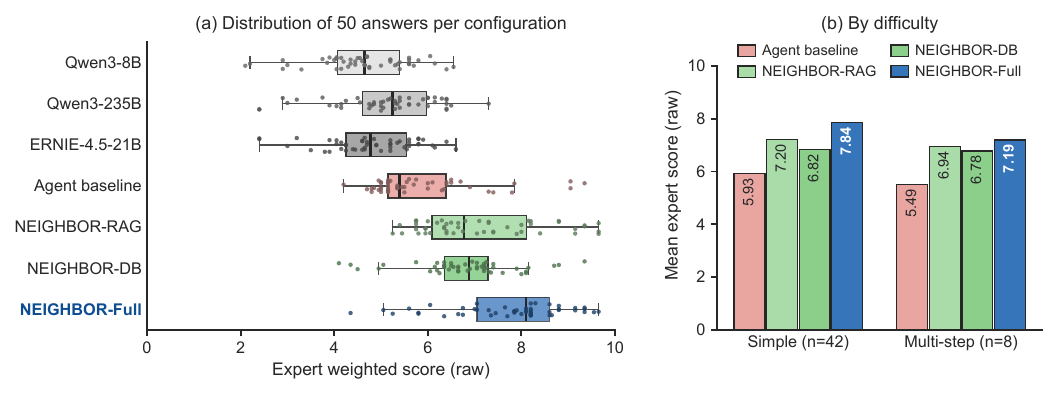}
\Description{Box plots of expert scores for 50 answers per configuration, and grouped bars of mean scores for simple and multi-step queries for the agent baseline and the three NEIGHBOR variants.}
\caption{Robustness of the gain. (a)~Expert scores of 50 answers per configuration (boxes: quartiles; points: individual answers); NEIGHBOR-Full shifts the whole distribution upward. (b)~Mean expert scores by difficulty; NEIGHBOR-Full ranks first in both strata.}
\label{fig:distribution}
\end{figure*}

The whole score distribution moves upward (Figure~\ref{fig:distribution}a). For NEIGHBOR-Full, the median is 8.10, 78\% of answers score 7 or higher (agent baseline: 14\%), and 49 of 50 answers score 5 or higher. The advantage also holds across difficulty (Figure~\ref{fig:distribution}b). NEIGHBOR-Full leads the agent baseline by 1.91 points on the 42 simple-retrieval queries (7.84 vs.\ 5.93) and by 1.70 on the eight multi-step queries (7.19 vs.\ 5.49), and it ranks first in both strata.

\subsection{Evidence-Aware Assessment of Tool-Grounded Answers}
\label{sec:judge}
\begin{figure*}[tbp]
\centering
\includegraphics[width=.84\textwidth]{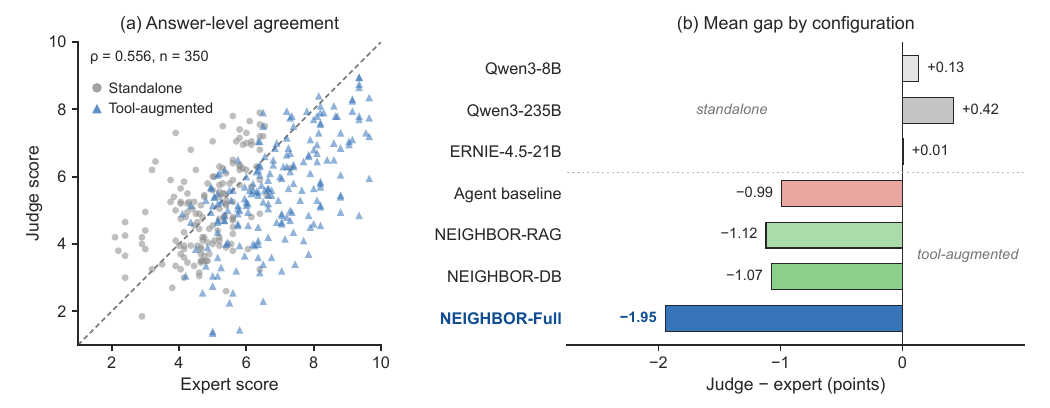}
\Description{Left: scatter plot of judge against expert scores for 350 answers, coloured by standalone or tool-augmented configuration. Right: bars of the mean judge-minus-expert gap for each configuration.}
\caption{LLM judge and domain expert on the same 350 answers. (a)~Answer-level scores; tool-augmented answers fall below the diagonal. (b)~Mean judge-minus-expert gap per configuration, negative by about 1--2 points for tool-augmented configurations and near zero for standalone models.}
\label{fig:agreement}
\end{figure*}

On the same 350 answers, the LLM judge tracks the expert at the answer level (pooled Spearman $\rho=0.556$) and ranks the seven configurations much as the expert does (Spearman $\rho=0.93$ over the seven means; Figure~\ref{fig:agreement}, Table~\ref{tab:agreement}). The judge's means lie within 0.42 points of the expert's for the three standalone models, and the differences concentrate on the four tool-augmented configurations, which the judge scores 0.99 to 1.95 points lower. By dimension, the tool-augmented gap is largest in hallucination control, at $-2.29$ points (standalone answers: $-0.68$).

\begin{table}[tbp]
\centering\small\setlength{\tabcolsep}{3.5pt}
\caption{Judge--expert agreement by dimension: answer-level Spearman $\rho$ and mean judge-minus-expert gap for standalone and tool-augmented answers.}
\label{tab:agreement}
\begin{tabular}{@{}lrrr@{}}
\toprule
Dimension & $\rho$ & Standalone & Tool-augmented \\
\midrule
Empathy & 0.218 & $+0.08$ & $-1.13$ \\
Utility & 0.665 & $+0.68$ & $-0.74$ \\
Compliance & 0.258 & $+0.41$ & $-1.14$ \\
Hallucination control & 0.581 & $-0.68$ & $-2.29$ \\
Conciseness & 0.350 & $+0.43$ & $-1.00$ \\
Retrieval & 0.654 & $+0.65$ & $-0.80$ \\
\midrule
Weighted score & 0.556 & $+0.19$ & $-1.28$ \\
\bottomrule
\end{tabular}
\end{table}

The rubrics' rules for admitting evidence account for this pattern. With the same inputs, dimensions, and weights (Section~\ref{sec:evaluation}), the expert's rubric admits specifics that the reference answer does not contain when they do not contradict it and reads citation markers as pointers to retrieved passages; the judge's rubric counts such specifics as fabricated. Tool-grounded answers carry many of them, such as an office's opening hours or the period of a record. In one question about a regional fee regulation, the reference answer defers the rates to a separately issued pricing policy, and all three NEIGHBOR variants state the same per-square-metre rates; the expert scored hallucination control 8 for each of these answers, and the judge scored 2 or 3 and recorded the rates as fabricated amounts. Pooled over all 350 answers, the judge's hallucination-control scores are 1.60 points below the expert's, the largest gap of the six dimensions. Choosing a configuration by the judge alone would select the document-only variant (judge mean 6.04) ahead of the full system (5.80), which the expert ranks first. Evaluating tool-using service agents thus calls for explicit rules on evidence beyond the reference answer, applied by an evidence-aware judge or by expert review.

\section{Designing for Agency in Resident-Facing Agents}
\label{sec:design}
Residents act on what a service assistant tells them: they pay a balance, visit an office, or wait for staff. In this setting, agency rests on answers a resident can verify and on a visible division of decisions between resident, assistant, and staff. Table~\ref{tab:design} turns the results of this study into five design principles. Each pairs a mechanism with the evidence from this study and the human--AI interaction guideline it implements~\cite{amershi2019guidelines,horvitz1999mixed}.

\begin{table*}[t]
\centering\small
\caption{Design principles for agency-preserving resident agents, each with the guideline it implements. G numbers refer to the guidelines for human--AI interaction of Amershi et al.~\cite{amershi2019guidelines}.}
\label{tab:design}
\begin{tabularx}{\textwidth}{@{}>{\raggedright\arraybackslash}p{0.25\textwidth}>{\raggedright\arraybackslash}X>{\raggedright\arraybackslash}X@{}}
\toprule
Principle (guideline) & Mechanism & Evidence in this study \\
\midrule
Checkable evidence (G9,~G11) & The synthesiser names the record, guide, and period behind each fact & Retrieval $+2.60$, hallucination control $+1.18$; no red-line flag on 200 tool-augmented answers \\
Commitments within authority (G1,~G10) & Commitment-control and referral targets; read-only access to the resident's own records & Compliance $+1.52$ with the same tools \\
Acknowledgement before procedure (G5) & Acknowledgement and procedural-clarity targets & Empathy $+2.52$, utility $+2.76$ \\
Shared initiative (mixed initiative~\cite{horvitz1999mixed}) & Two suggested follow-ups; the resident chooses & Deployed interface (Figure~\ref{fig:architecture}b) \\
Explicit evidence rules for assessment (LLM-as-judge~\cite{zheng2023judging}) & The expert's rubric admits specifics that do not contradict the reference answer & Judge $-1.60$ on hallucination control across 350 answers \\
\bottomrule
\end{tabularx}
\end{table*}

\paragraph{Checkable evidence.} The synthesiser names the record or guide behind each fact and, for a figure, the period it covers. A resident who doubts a balance can compare it with their own bill and contest it through the named office (G9), and the citation shows where the answer comes from (G11). An interface can show the cited record beside the answer and link it to the correction channel.

\paragraph{Commitments within the assistant's authority.} Acceptance, corrections, refunds, and repairs need a person with authority. The commitment-control and human-referral targets train the model to state what the assistant can do and to pass authority-bound requests to the responsible office, so the resident learns early which step needs staff (G1, G10). Read-only access to the resident's own records (Section~\ref{sec:system}) keeps the same boundary at the data layer, and in the interface a referral names a concrete office and channel.

\paragraph{Acknowledgement before procedure.} The two largest gains, in utility and empathy, correspond to the targets for procedural clarity and acknowledgement. Residents receive recognition of their concern followed by steps they can carry out, the conduct expected at a service desk (G5). Because this conduct sits in the adapted model, every role that calls the model shares it.

\paragraph{Shared initiative.} After each answer, the follow-up predictor proposes two related questions, and the resident decides whether to ask one (Figure~\ref{fig:architecture}b). Initiative in anticipating information needs thus lies with the assistant, while the resident steers the conversation and staff keep the decisions that need authority, the division that mixed-initiative design describes~\cite{horvitz1999mixed}.

\paragraph{Explicit evidence rules for assessment.} An assessor who scores against a reference answer needs a rule for specifics that the reference does not contain, and tool-grounded answers supply many of them. In this study, the rule changed which configuration ranked first (Section~\ref{sec:judge}). Teams that select configurations with an automated judge can write the rule into the judge's rubric, or supply the retrieved passages so that such specifics can be checked, and treat citation markers as pointers to those passages.

\section{Conclusion}
This paper has presented NEIGHBOR, a deployed agent for community resident services that trains service behaviour into a LoRA-adapted Qwen3-235B-A22B and runs the adapted model over document retrieval and schema-checked database access. With the tools held fixed, fine-tuning raised the independent expert's score from 5.86 to 7.74/10 and improved every rated dimension, and no NEIGHBOR answer received a red-line flag. Service conduct that residents depend on, such as naming the record behind a figure and leaving decisions to staff, can therefore be trained into the same model that uses the tools. The judge--expert comparison adds that assessing such answers requires a rubric that states how specifics beyond the reference answer are treated.

\section*{Data and Code Availability}
The anonymous per-question rating vectors, the statistics script, and the plotting code accompany the source files of this preprint. Machine-translated English versions of 500 de-identified examples from the SFT data, none of them from the test queries, are available at \url{https://github.com/tongjiu123/NEIGHBOR} together with the scoring rubric and the training documentation; further data will follow as de-identification proceeds. Examples in the paper replace community, office, staff, and account identifiers, and their de-identification extends to contextual combinations~\cite{ico2025anonymisation}. The full instruction corpus and evaluation dialogues remain under the reuse conditions of their source materials.